\documentclass[runningheads]{llncs}
\usepackage[T1]{fontenc}
\usepackage{graphicx}
\usepackage{booktabs}
\usepackage[misc]{ifsym}

\usepackage{subcaption}
\usepackage{tikz} 
\usetikzlibrary{calc}
\usepackage{float}
\usepackage{multirow}
\usepackage{algorithm}
\usepackage{algpseudocode}
\usepackage{tabularx}
\usepackage{amsmath}
\usepackage{amssymb}

\usepackage{hyperref}
\usepackage{url} 
\usepackage{cleveref}
\usepackage{textcomp}
\usepackage{empheq}
\usepackage{siunitx}

\newcommand{\argmin}[1]{\underset{#1}{\operatorname{arg}\,\operatorname{min}}\;}
\newcommand{\argmax}[1]{\underset{#1}{\operatorname{arg}\,\operatorname{max}}\;}

\DeclareMathOperator{\Neighbors}{Neighbors}
\DeclareMathOperator{\cut}{Cut}
\DeclareMathOperator{\Leiden}{Leiden}
\DeclareMathOperator{\size}{size}

\newcommand{\ie}{\textit{i}.\textit{e}. }
\newcommand{\eg}{\textit{e}.\textit{g}., }

\begin{document}

\title{Leiden-Fusion Partitioning Method for Effective Distributed Training of Graph Embeddings}

\titlerunning{Leiden-Fusion Graph Partitioning Method}
\author{Yuhe Bai\and
Camelia Constantin\and
Hubert Naacke(\Letter) }

\authorrunning{Yuhe Bai et al.}

\institute{
LIP6, Sorbonne University, Paris, France \\
\email{\{firstname.lastname\}@lip6.fr}}

\maketitle

\begin{abstract}
In the area of large-scale training of graph embeddings, effective training frameworks and partitioning methods are critical for handling large networks. However, they face two major challenges: 1) existing synchronized distributed frameworks require continuous communication to access information from other machines, and 2) the inability of current partitioning methods to ensure that subgraphs remain connected components without isolated nodes, which is essential for effective training of GNNs since training relies on information aggregation from neighboring nodes. To address these issues, we introduce a novel partitioning method, named Leiden-Fusion, designed for large-scale training of graphs with minimal communication.
Our method extends the Leiden community detection algorithm with a greedy algorithm that merges the smallest communities with highly connected neighboring communities. 
Our method guarantees that, for an initially connected graph, each partition is a densely connected subgraph with no isolated nodes. 
After obtaining the partitions, we train a GNN for each partition independently, and finally integrate all embeddings for node classification tasks, which significantly reduces the need for network communication and enhances the efficiency of distributed graph training. We demonstrate the effectiveness of our method through extensive evaluations on several benchmark datasets, achieving high efficiency while preserving the quality of the graph embeddings for node classification tasks.

\end{abstract}

\section{Introduction}
Graph embeddings have become a fundamental technique in machine learning, providing a powerful means of dealing with complex structured data. By transforming nodes, edges, and their interactions within a graph into a compact, lower-dimensional vector space, graph embeddings allow machine learning techniques to be applied to graph data with increased efficiency.

To compute graph embeddings, Graph Neural Networks (GNNs) have gained prominence due to their ability to exploit the inherent structure of graph data. Among them, the most popular are Graph Convolutional Networks (GCN)\cite{gcn} and GraphSAGE\cite{sage}. Using graph convolution operations, GNNs iteratively aggregate and transform the embeddings of neighboring nodes, culminating in a representation that captures both local and global graph structures.

However, the scalability of GNNs to very large graphs presents a significant challenge. While parallel processing can enhance the efficiency of GNNs by allocating computations across multiple processors or GPUs, for extremely large graphs that exceed the capacity of a single machine, it is crucial to partition the graph and distribute the computational load across multiple machines. While traditional partitioning approaches facilitate distributed learning, they often fail to preserve the structural coherence of the original graph. They typically generate subgraphs that contain multiple connected components and isolated nodes, undermining the performance of GNNs. A connected component is a subgraph in which every pair of nodes is connected by a
path, and an isolated node represents a vertex of a graph with no edges and thus of a degree zero. This is because the effectiveness of GNNs depends on the premise that a node's embedding is enriched by the embeddings of its neighbors; if these neighbors lie outside the subgraph, not only will there be more communication, but the quality of the embeddings will also decrease.

To address these challenges, our work introduces a novel partitioning method designed to preserve the structural integrity of subgraphs in a distributed learning framework, followed by a local training strategy. Specifically, we ensure that for any given graph that initially consists of a single connected component, each partition remains a connected component with no isolated nodes. This not only preserves the contextual relevance of node embeddings but also allows local training and eliminates the need for inter-subgraph communication, thereby increasing the efficiency of distributed GNN training. Our contributions are as follows:

\begin{enumerate}
\item For an initially connected graph, we proposed a novel partitioning method that guarantees the structural integrity of subgraphs by ensuring that each subgraph remains a single connected component with no isolated nodes.
\item  By using single connected components as partitions, we demonstrate the feasibility of achieving high training efficiency for GNNs without sacrificing much accuracy, paving the way for more scalable and efficient distributed learning on very large graphs.
\end{enumerate}

The paper is organized as follows: \Cref{sec:background} presents background knowledge about GNNs and graph embeddings, related work is presented in~\Cref{sec:related_work}. \Cref{sec:method} presents our novel Leiden-Fusion algorithm, and experimental results are discussed in \Cref{sec:experiments}.

\section{Background on Graph Embeddings}
\label{sec:background}
Graph Neural Networks (GNNs) extend neural network methods to graph data. A typical GNN layer updates the representation of a node based on its neighbors. Graph Convolutional Networks (GCN)~\cite{gcn} and GraphSAGE~\cite{sage} represent two major advances in the field of GNNs, each introducing unique strategies for aggregating neighborhood information to improve node embeddings. The resulting embeddings are critical in a variety of applications, including but not limited to node classification\cite{node_class}, question answering\cite{qa1}, and recommender systems\cite{recom1}.

\textbf{GCN\cite{gcn}:} 
The key idea behind GCN is to update the representation of a node by aggregating the representations of its neighbors. This approach captures the local graph topology in the node embeddings. The formula given for GCN is:

\begin{equation}
\mathbf{h}_v^{l}=\sigma\left(\frac{1}{|N(v)|}\sum_{u \in N(v)} \mathbf{W}^l 
\mathbf{h}_u^{l-1}\right)
\end{equation}

This formula represents how the representation $\mathbf{h}_v^{l} $ of a node $v$ at layer $l$ is updated. It does this by applying a nonlinear activation function $\sigma$ (\eg ReLU function) to the normalized sum of the representations of its $u$ ($u \in N(v)$) neighbors from the previous layer $\mathbf{h}_u^{l-1}$.
$\mathbf{W}^{l}$ is the weight matrix for the layer $l$.

\textbf{GraphSAGE\cite{sage}:} GraphSAGE extends the idea of GCN by incorporating the node's own features along with its neighbors, and by using a sampling strategy that selects a fixed subset of neighbors to aggregate information from, allowing scalability in large graph settings. The formula for GraphSAGE is:
\begin{equation}
\mathbf{h}_v^{l}=\sigma\left(\mathbf{W}^{l} \cdot \operatorname{CONCAT}\left(\mathbf{h}_v^{l-1}, \operatorname{AGG}\left(\left\{\mathbf{h}_u^{l-1}, \forall u \in N(v)\right\}\right)\right)\right)
\end{equation}

In this equation, the new representation of a node \(v\) at layer \(l\) is obtained by first concatenating the representation of its previous layer $\mathbf{h}_v^{l-1}$ with an aggregated representation of its sampled neighbors' features $\mathbf{h}_u^{l-1}$. The aggregation (\(\operatorname{AGG}\)) can be a mean, sum, or max operation. 

This method allows for efficient computation on large-scale graphs and enriches the node embeddings with both central node and sampled neighborhood information.

Thus, the effectiveness of these models relies heavily on their ability to aggregate information from neighboring nodes, underscoring the importance of a partitioning method in a distributed setting that computes partitions as connected graph components. 
Our partitioning method ensures that the structural integrity of the graph is maintained within each partition, which is crucial for effective local model training.

\begin{figure}[ht]
  \centering
  \includegraphics[width=0.9\textwidth]{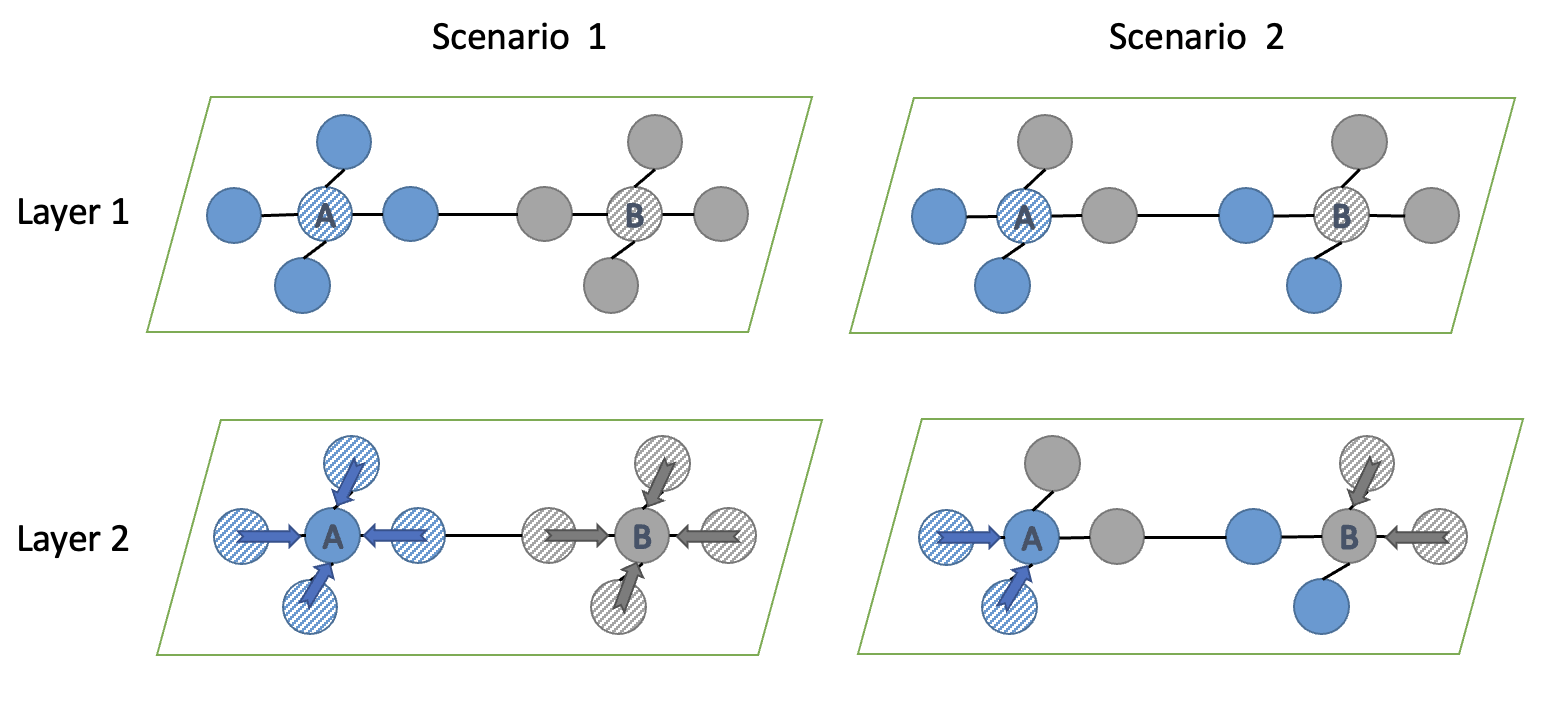}\hfill
  \caption{Aggregation of nodes A and B in GNNs with different partitioning strategies}
  \label{fig:Aggregation}
\end{figure}

\Cref{fig:Aggregation} visually illustrates the process of neighbor aggregation for nodes A and B, contrasting two scenarios based on the partitioning of the graph into subgraphs colored blue (first partition) and gray (second partition). 

On the left, both partitions contain a single connected component, ensuring that full neighbor information is available for aggregation. 
On the right, however, the presence of multiple components and isolated nodes within each partition severely limits the information that nodes A and B can aggregate. Nodes A and B can only aggregate two neighbors instead of four.

More specifically, in a distributed framework with no communication, a lot of neighbor information is lost; with synchronization, a lot of communication occurs and there is a delay in information transformation. This illustrates the impact of graph connectivity on the update process in GNN layers.

This illustration highlights the importance of ensuring that each subgraph not only remains a connected component but also avoids isolated nodes to maximize the effectiveness of distributed GNN training. Our partitioning algorithm is specifically designed with this goal in mind and aims to improve the efficiency and effectiveness of GNNs in distributed environments.

\section{Related Work}
\label{sec:related_work}

\subsection{Partitioning Methods}

The goal of most partitioning methods is to reduce edge cuts and ensure load balance, to reduce the communication of synchronized distributed frameworks. We will introduce some of the SOTA partitioning methods.

\textbf{METIS}\cite{metis} is one of the most popular algorithms and is used by most SOTA distributed frameworks. For each machine, it aims to form a diagonal-like block in the adjacency matrix, so that when a trainer processes samples in the local partition, most of the embeddings accessed by the batch fall in the local partition, and thus there is little network communication for accessing entity embeddings. METIS focuses mainly on balancing the node size of the partition and minimizing edge cuts.

However, this approach does not directly focus on the component structure within the partitions, which means that it may split a component into multiple partitions, resulting in many isolated nodes. This is problematic for GNN models, which, as discussed earlier, rely heavily on the integrity of the graph structure for effective training.

\textbf{LPA:} The Label Propagation Algorithm\cite{spinner}\cite{pregel}\cite{spark} (LPA) was originally designed to detect communities in graphs, using the network structure to determine the communities. In LPA, each node in the graph is initially assigned a unique label. At each iteration of the algorithm, nodes adopt the label that most of their neighbors currently have. This update rule can be written as:

\begin{equation}
\text{label}(v) \leftarrow \text{mode}(\{\text{label}(u) : u \in \mathcal{N}(v)\})
\end{equation}

Where \( \text{label}(v) \) is the label of the node \( v \), and \( \mathcal{N}(v) \) is the set of neighbors of \( v \). The mode function selects the most frequent label among the neighbors of a node. The algorithm runs iteratively until convergence or a certain number of epochs is reached, at which point nodes with the same label are considered to be in the same community. One of the main advantages of LPA is its ability to scale naturally to large networks due to its simplicity. To use it for graph partitioning in distributed learning, each node is initially assigned a label ranging from 0 to the number of partitions $K$.

However, the algorithm has several limitations. It can be sensitive to the initial label assignment and can produce different results on different runs. It can also converge to a trivial solution where all nodes end up with the same label in highly connected graphs. In LPA, each node is initially randomly assigned a label from 0 to n. This means that, for example, for label 0, there may initially be some nodes with label 0 at different positions in the graph. They then propagate separately, forming many small components centered on themselves, resulting in partition 0 having many components quite far apart from each other. 

Many other partitioning methods are also used to address specific needs, such as random partitioning, a simple approach where nodes (or edges) are randomly assigned to partitions. It can provide load balancing and high diversity within a partition, but in synchronized frameworks, the communication overhead can be very high; in unsynchronized frameworks, it can lead to poor quality embeddings because each node loses most of its neighbors' information.

\subsection{Distributed Training Frameworks}

Many frameworks have emerged to facilitate efficient and scalable distributed training of graph embeddings. The key to optimizing distributed training is to reduce the communication required to retrieve and update embeddings. However, no matter how it is reduced, most existing techniques for distributed graph embedding, such as Deep Graph Library (DGL)\cite{dgl} and PyTorch BigGraph (PBG)\cite{pbg}, require continuous communication.

Spark Local\cite{spark} is one of the first frameworks to perform local training of subgraphs to avoid continuous communication. It partitions a graph into subgraphs using LPA (Label Propagation Algorithm) while considering a "landmark graph" which is a small subset of the graph based on node degrees, then they put the landmarks into each subgraph, learn their embeddings locally, and reconcile the embedding spaces using SVD (Singular Value Decomposition) based on the landmark embeddings. However, the quality of the embeddings is degraded because the LPA algorithm can lead to poor-quality partitions. In addition, it is very time-consuming to find the landmarks and add the edges connecting them to each partition.

To address the shortcomings of the current partitioning and distributed training methods, we will introduce our Leiden-Fusion method in the next section.

\section{Leiden-Fusion Method}
\label{sec:method}

In this section, we outline the main contributions of our work. First, we define the essential features of partitions that allow high-quality embeddings to be computed independently on each partition. We then present a detailed description of our two-step approach.

\subsection{Essential Features for Graph Partitioning}

As we discussed earlier, we assume that for local training of GNN on subgraphs to be effective, the following conditions must be met: 

\begin{enumerate}
\item \textbf{Each partition should contain one densely connected component.} By ensuring this, most nodes can retain all neighbor information. Only for boundary nodes, a small amount of neighbor information will be lost.

\item \textbf{There should be no isolated nodes.} Similar to ensuring one densely connected component, if there are isolated nodes in the subgraphs, these nodes will have no neighbors to aggregate with to update their information, leading to poor training results.

\end{enumerate}

Existing partitioning methods cannot meet these two requirements as we discussed in~\Cref{sec:related_work}. Our partitioning method is designed to meet these requirements. The main idea of our approach is to rely on a community detection algorithm and then merge communities in a way that results in densely connected partitions free of isolated nodes.

\subsection{Leiden Community Detection}
The first step is to obtain densely connected communities using the Leiden algorithm:

The Leiden algorithm~\cite{leiden} is an iterative community detection method that improves on the well-known Louvain algorithm~\cite{louvain}, with improvements in terms of quality and speed. The primary goal of the Leiden algorithm is to optimize a modularity function:
\begin{equation}
Q = \frac{1}{2m} \sum_{c} \left( e_{c} - \gamma\frac{K_c^2}{2m} \right)
\end{equation}
Where $e_c$ is the actual number of edges in the community $c$. The expected number of edges is $\frac{K_c^2}{2m}$, where $K_c$ is the sum of the degrees of the nodes in community $c$ and $m$ is the total number of edges in the network. This modularity is a scalar value that measures the density of links inside communities compared to links between communities.
By maximizing the modularity function, Leiden ensures that the resulting communities are densely connected. We abstract the Leiden community detection in \Cref{def:community}.
\begin{definition}[Leiden communities]
\label{def:community}
Let  G = $(V,E)$ be a graph and $C = \{C_1, \dots, C_n\}$ be a partition of $V$ which implies $C_i \cap C_j = \emptyset$ for $i \neq j$. Let $G_i$ be the projection of G onto $C_i$.
Let $S$ be the maximum expected size of a community.
$\textbf{Leiden}: G \mapsto C $  associates $G$ with $C$ communities such that it maximizes the modularity of the communities, and each community has less than $S$ vertices, \ie
$\forall C_i \in C, |G_i| \leq S $.
\end{definition}

\subsection{Community Fusion}
Since the number of communities obtained by Leiden is, in most cases, much larger than the expected number of partitions $k$, which typically corresponds to the number of machines in a distributed training environment. To address this issue, we propose a novel fusion method to merge these communities.

Our solution is based on the notions of edge cut, defined in \Cref{def:edge_cut}, and community neighborhood, defined in \Cref{def:neighbor_communities}.

\begin{definition}[Edge cut]
\label{def:edge_cut}
Let $G=(V,E)$ be a graph. Let $V_i$, $V_j$ be two disjoint subsets of $V$. 
Let $G_i$ (resp. $G_j$) be the projection of G on $V_i$ (resp. $V_j$).
We define $Cut(G_i, G_j)$ as the set of edges connecting $G_i$ with $G_j$. We have:
$Cut(G_i, G_j) = \{(v, v') \in E| v \in G_i \wedge  v' \in G_j \}$
\end{definition}

\begin{definition}[Neighbor communities]
\label{def:neighbor_communities}
Let $C$ be a set of communities in the graph $G$. The neighboring communities, denoted $Neighbors(C_i)$, are the set of communities that are adjacent to $C_i$,  \ie 
$Neighbors(C_i) = \{ C_j \in C |  Cut(C_i, C_j) \neq \emptyset\}$
\end{definition}

\begin{figure}[H]
  \centering
  \begin{subfigure}[b]{0.3\linewidth}
    \includegraphics[width=0.9\linewidth]{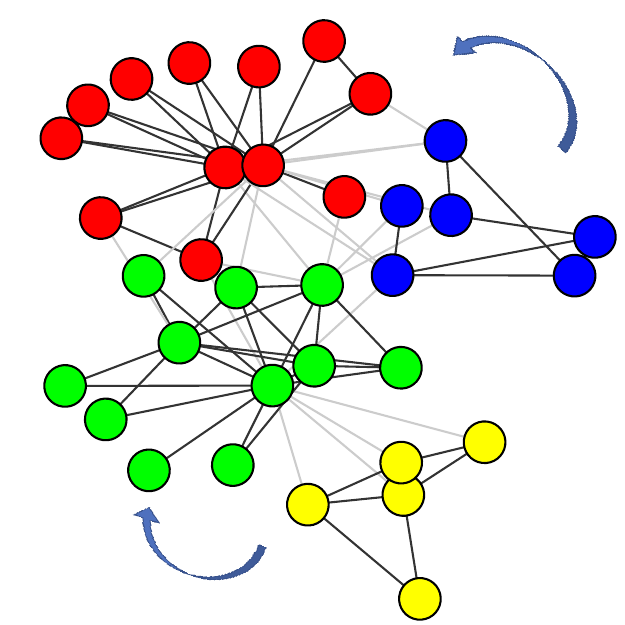}
    \caption{Leiden communities}
    \label{fig:leiden}
  \end{subfigure}
  \hspace{0.5cm}
  \begin{subfigure}[b]{0.3\linewidth}
    \includegraphics[width=0.9\linewidth]{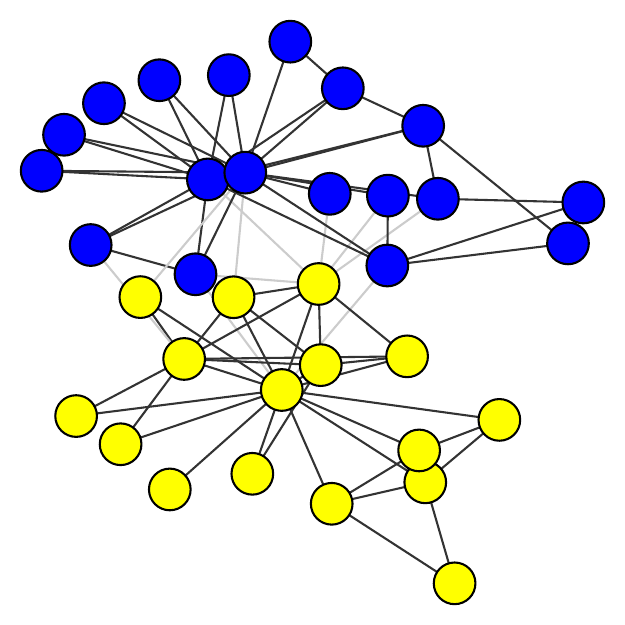}
    \caption{Our partitions}
    \label{fig:fusion}
  \end{subfigure}
  \caption{Visualization of Leiden community detection and fusion process}
  \label{fig:partitioning}
\end{figure}

Starting from the initial partitions computed by the Leiden algorithm, for a given partitioning number $k$, our method iteratively computes $k$ balanced partitions by merging existing partitions with their neighbors. The intuition of the Leiden-Fusion algorithm is shown on Zachary’s karate club network\cite{karate} in ~\Cref{fig:partitioning}. The goal is to partition the Karate graph into two partitions. First, we get 4 communities through the Leiden community detection algorithm, and then we start from the smallest community, which is the yellow community. We find its most connected neighbor, which is the green community, and merge them. Then the blue community becomes the smallest one to merge with the red one, and finally we get 2 partitions.

The Leiden-Fusion algorithm is described in \Cref{alg:LF}. The parameters $\alpha$ and $\beta$ are used to control the number of nodes assigned to each partition and the maximum size of the initial communities computed by Leiden. We aim to compute balanced partitions whose size is controlled by the variable {\em max\_part\_size}, with a tolerance threshold given by $\alpha$ (line 3). We first apply the Leiden community detection algorithm to identify numerous small communities $C$ within the graph $G$ (line 4). Communities are iteratively merged to form larger and larger communities, with each iteration selecting the smallest community in terms of number of nodes ($c_{min}$) and gradually merging it with its largest edge-cut neighbor community ($c_{max\_cut}$) (lines 5-10). The fusion process ends when $|C|$ equals the desired number of partitions $k$.

\begin{algorithm}[H]
\caption{Leiden-Fusion Partitioning Algorithm}
\label{alg:LF}

\begin{algorithmic}[1]
\State \textbf{Input:} $G$: graph, $k$: number of partitions, $\alpha$, $\beta$
\State \textbf{Output:} $C$ composed of k subgraphs
\State $max\_part\_size \gets \frac{size(G)}{k} \times (1+\alpha)$ 
\State $C \gets \Leiden(G,\beta \times max\_part\_size )$ \quad  // C is a set of subgraphs

\While{|C| $> k$}
    \State $c_{min} \gets \argmin{c \in C} size(c)$ \quad // get the smallest community 
    \State $c_{max\_cut} \gets \operatorname{LargestEdgeCutNeighbor}(c_{min}, max\_part\_size)$
    \State $c_{merged} \gets c_{max\_cut}\cup c_{min}  $ \quad // merge graph $c_{min}$ with graph $c_{max\_cut}$
    \State $C \gets (C \setminus \{c_{min}, c_{max\_cut} \}) \cup  \{ c_{merged}\} $ \quad // update communities  
\EndWhile
\State \textbf{return} C 
\end{algorithmic}
\end{algorithm}
The largest edge-cut neighboring community is computed by~\Cref{alg:LF2}. For each community $v$ to be merged, it finds the most connected community $c$ (given by $|\cut(v,c)|$, which is the number of edges between $v$ and $c$) within the size limit given by {\em max\_part\_size} (lines 3-5). If for every neighbor community $c$ the merge exceeds the size limit {\em max\_part\_size}, $v$ will be merged with its smallest neighbor to ensure load balance (lines 6-8).

\begin{algorithm}
\caption{LargestEdgeCutNeighbor}
\label{alg:LF2}
\begin{algorithmic}[1]

\State \textbf{Input:} $v, max\_part\_size$
\State \textbf{Output:} $u$
\State $ N \gets  \{ c \in \Neighbors(v) | \size(c) + \size(v) <  max\_part\_size \} $

\If{$N \neq \emptyset$}
    \State $u \gets \argmax{c \in N} |\cut(v, c)| $ ~ // get the most connected neighbor among $N$
\Else
\State $u\gets \argmin{c \in \Neighbors(v)}\size(c)$ ~ // get the smallest neighbor
\EndIf
\State \Return $u$
\end{algorithmic}
\end{algorithm}

Each partition obtained by this method consists of a single unified component since the initial graph is a connected component and each community computed by the Leiden algorithm is densely connected without isolated nodes.

\subsection{Partition Visualization on Karate Dataset}
\begin{figure}[H]
    \centering
    \begin{subfigure}[b]{0.22\linewidth}
        \centering
        \includegraphics[width=\linewidth]{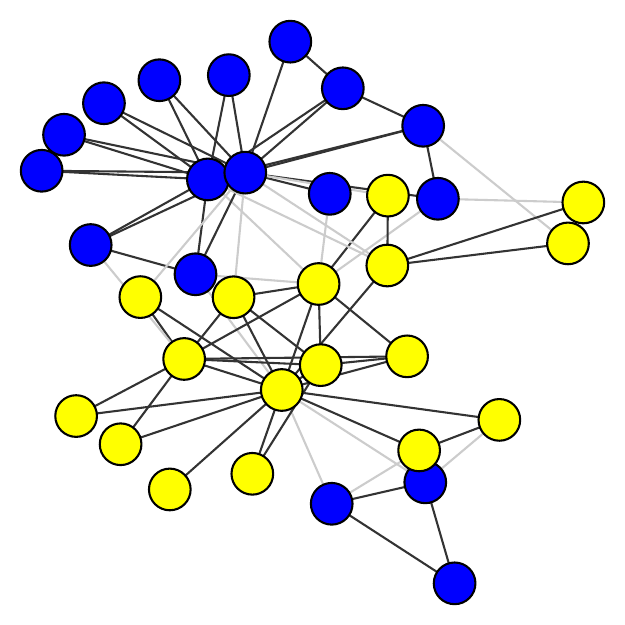}
        \caption{LPA}
        \label{fig:lpa}
    \end{subfigure}
    \begin{subfigure}[b]{0.22\linewidth}
        \centering
        \includegraphics[width=\linewidth]{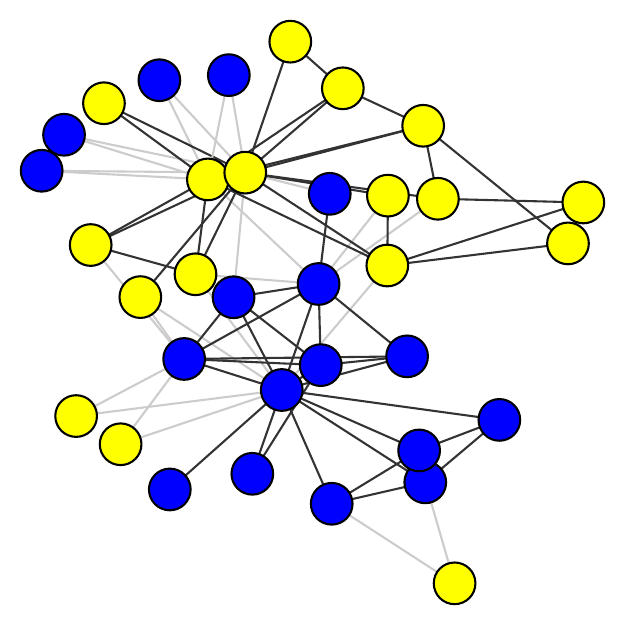}
        \caption{METIS}
        \label{fig:metis}
    \end{subfigure}
    \begin{subfigure}[b]{0.22\linewidth}
        \centering
        \includegraphics[width=\linewidth]{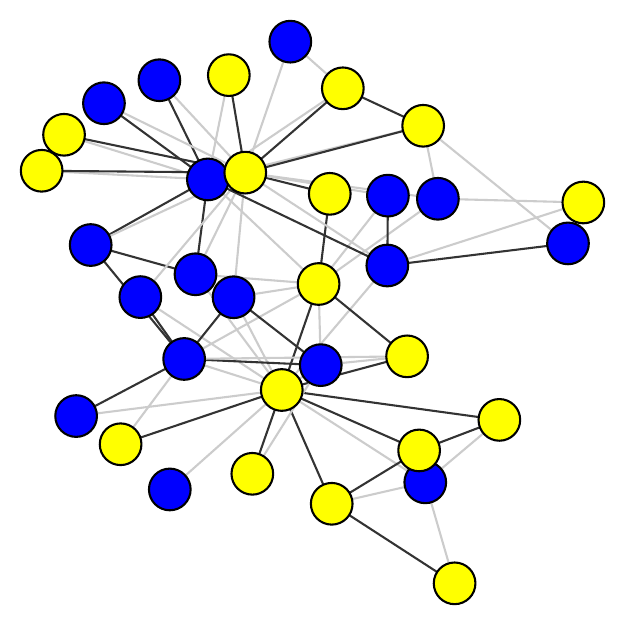}
        \caption{Random}
        \label{fig:random}
    \end{subfigure}
    \begin{subfigure}[b]{0.22\linewidth}
        \centering
        \includegraphics[width=\linewidth]{figures/ours.pdf}
        \caption{Ours}
        \label{fig:ours}
    \end{subfigure} 
    \caption{Comparison of partitioning methods on Karate dataset. \textcolor{blue}{$\bullet$} Partition 0 \textcolor{yellow}{$\bullet$} Partition 1}
    \label{fig:partition-comparison}
\end{figure}

To prove the effectiveness of our algorithm, we compared METIS, LPA, Random and our LF on this Karate dataset, the results are shown in~\Cref{fig:partition-comparison} and~\Cref{tab:partitioning_methods}. We can see that our algorithm outperforms on both criteria in the toy example. From \Cref{fig:partition-comparison} we can see that the LPA method can lead to poor quality partitions because it is sensitive to the initial label assignment. If two nodes at different positions in the graph are assigned the same label (partition 0 in this example), they may propagate to form many components at different positions in the graph, as shown in the figure. Similarly for METIS, we can see that there are many isolated nodes and many components in the partitions.

\vspace{-10pt}

\vspace{-20pt}

As can be seen from \Cref{tab:partitioning_methods}, in this toy example, the partitions obtained by our LF method have zero isolated nodes, each partition has only one component and minimal edge cuts.

\begin{table}[H]
\centering
\begin{tabular}{@{}lcccccc@{}}
\toprule
\multirow{2}{*}{Method} & \multicolumn{2}{c}{Isolated Nodes} & \multicolumn{2}{c}{Components} & {Edge Cuts} \\
\cmidrule(l){2-7}
& Part 0 & Part 1 & Part 0 & Part 1 & Part 0 \& 1 \\
\midrule
LPA & 0 & 0 & 2 & 1 & 17 \\
METIS & 4 & 3 & 5 & 4 & 25 \\
Random & 4 & 1 & 5 & 2 & 45 \\
Ours & 0 & 0 & \textbf{1} & \textbf{1} & \textbf{10} &\\
\bottomrule
\end{tabular}
\vspace{0.3cm}
\caption{Evaluation of Partitioning Methods on Karate Dataset}
\label{tab:partitioning_methods}
\end{table}

\vspace{-30pt}

\paragraph{Advantages of the Proposed Two-Step Method: } Our fusion method can be applied to any graph partitioning technique, but we chose the Leiden community detection method because of its ability to produce well-connected communities. However, Leiden communities vary in size and do not allow specifying the desired number of communities. Our fusion method addresses these limitations by allowing the generation of a specified number of balanced communities. Other graph partitioning methods, such as METIS and LPA, are designed to achieve a given number of partitions. However, they often produce multiple components and isolated nodes, making graph structure reconstruction time-consuming, as shown in the experimental section. This process involves identifying each component within a partition and treating them as separate partitions for fusion.

\section{Experimental Results}
\label{sec:experiments}

\textbf{Setup:} We first perform the partitioning methods on one CPU in a centralized way. For METIS, we used the library provided by DGL\cite{dgl}. For LPA, we reproduced the method of Spark Local\cite{spark}, and then we implemented our Leiden-Fusion method. 

Due to resource limitations, we ran the training process sequentially on a single machine for each partition, which is equivalent to a fully distributed implementation since there is no communication during the training process. 
The hardware used includes a DELL PowerEdge R650xs with 125 GB of memory and an Intel Xeon Silver 4310 processor with 24 cores / 48 threads @ 2.10 GHz, and a DELL PowerEdge R750xa with 2 TB of memory equipped with two Intel Xeon Gold 6330 CPUs, each with 56 cores / 112 threads @ 2.00 GHz, and four NVIDIA A100 80 GB PCIe GPUs. The code is available at \url{https://github.com/YuheBAI/leiden-fusion}.

\textbf{Datasets:} The datasets we used are the Arxiv and Proteins datasets for node prediction tasks from the Open Graph Benchmark (OGB)\cite{ogb}. The Arxiv dataset is a directed graph, representing the citation network between all Computer Science (CS) Arxiv papers indexed by MAG\cite{mag}. The graph contains \num{169343} nodes and \num{1166243} edges. The task is to predict the label of each node from 40 subject areas of Arxiv CS papers, which is a multi-classification task. The proteins\cite{proteins} dataset is an undirected, weighted, and typed (by species) graph. Nodes represent proteins, and edges indicate different types of biologically meaningful associations between proteins, such as physical interactions, co-expression, or homology\cite{proteins}\cite{proteins2}. The graph contains \num{132534} nodes and \num{39561252} edges. The task is to predict the presence of protein functions in a multi-label binary classification setup, where there are a total of 112 types of labels to predict. Performance is measured by the average of the ROC-AUC values over the 112 tasks. 

\textbf{Hyperparameter Settings:} In the experiments conducted for this paper, specific hyperparameters were set for different parts of the process. During the graph partitioning phase, $\alpha$, which controls the partition size, was set to 0.05, and $\beta$, which controls the size of the Leiden community, was set to 0.5. For the GNN training phase, we used the same hyperparameters as recommended by OGB\cite{ogb}, with the number of epochs reduced to 80 for the Arxiv dataset to avoid overfitting, since training was performed on smaller subgraphs.

\subsection{Analysis of Partitions}
To evaluate the effectiveness of our partitioning method in terms of subgraph quality, according to the literatures\cite{graphcomputingsurvey}\cite{load_balance}\cite{repli_factor}, we adopted the following metrics to measure subgraphs: 

\begin{figure}[htbp]
  \centering
  \begin{subfigure}[b]{0.32\linewidth}
    \includegraphics[width=\linewidth]{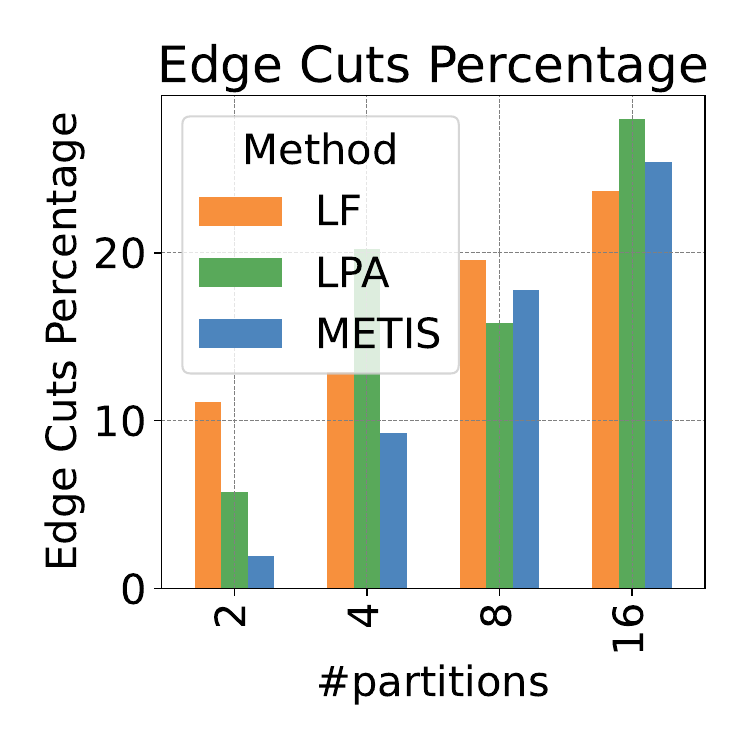}
    \label{fig:leiden}
  \end{subfigure}
  \begin{subfigure}[b]{0.32\linewidth}
    \includegraphics[width=\linewidth]{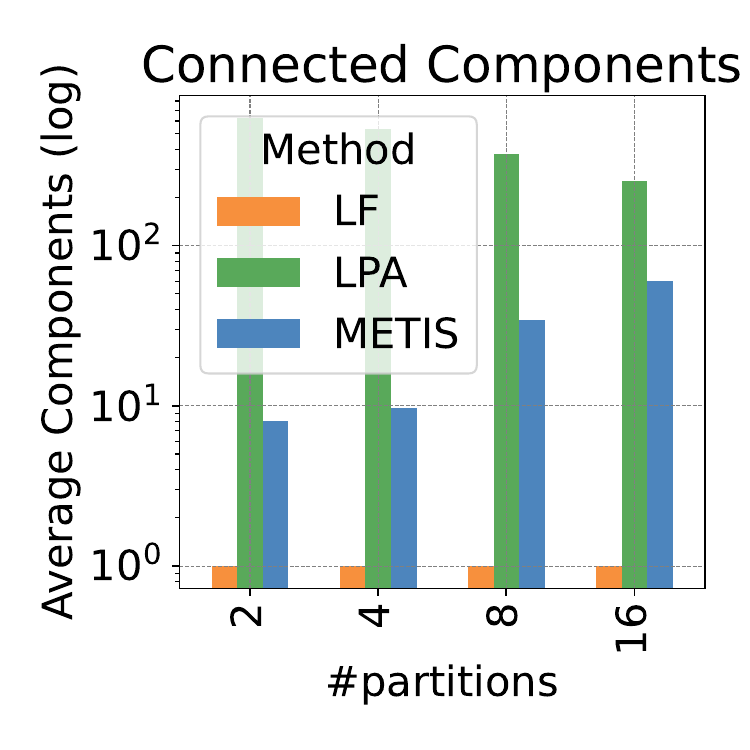}
    \label{fig:fusion}
  \end{subfigure}
  \begin{subfigure}[b]{0.32\linewidth}
    \includegraphics[width=\linewidth]{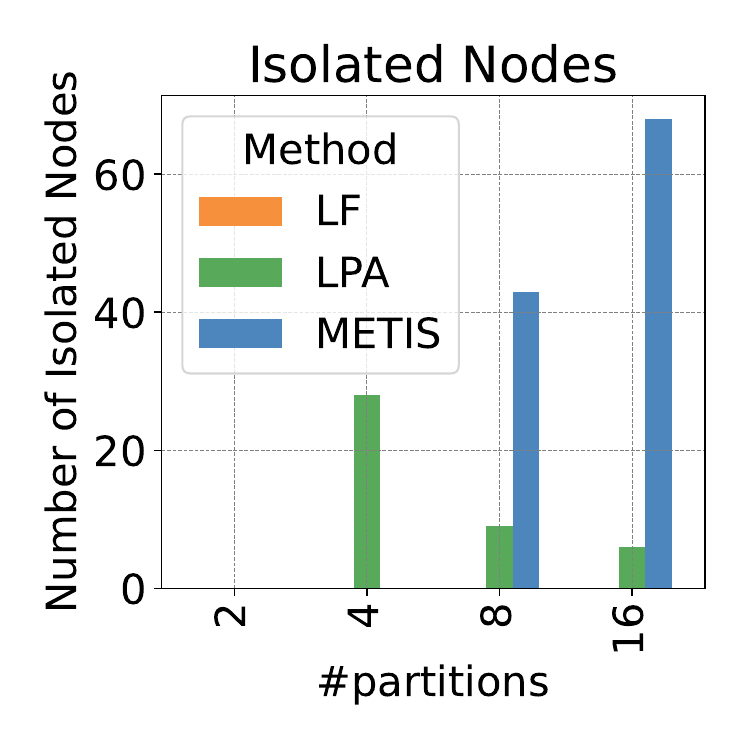}
    \label{fig:fusion}
  \end{subfigure}
  \vspace{-0.9cm}
%
  \begin{subfigure}[b]{0.32\linewidth}
    \includegraphics[width=\linewidth]{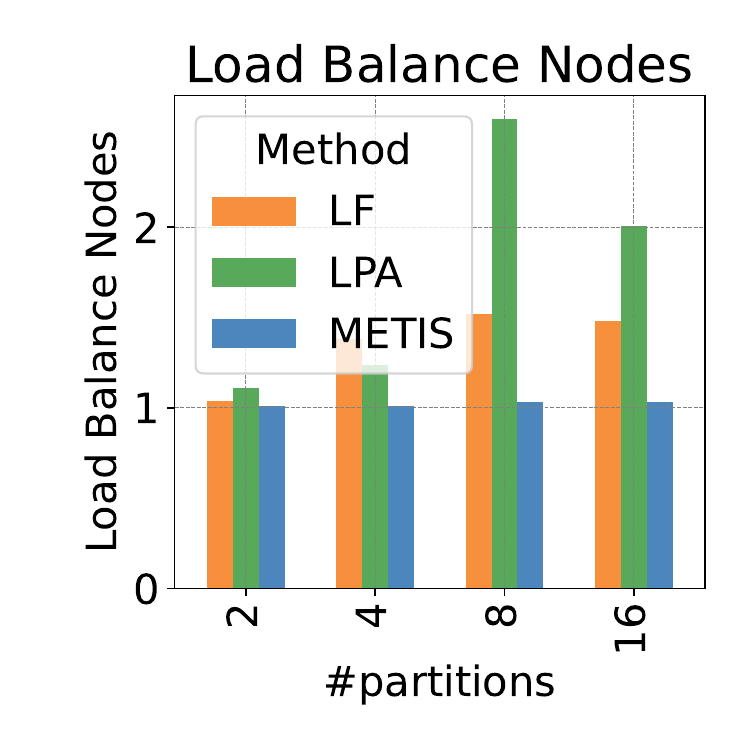}
    \label{fig:leiden}
  \end{subfigure}
  \begin{subfigure}[b]{0.32\linewidth}
    \includegraphics[width=\linewidth]{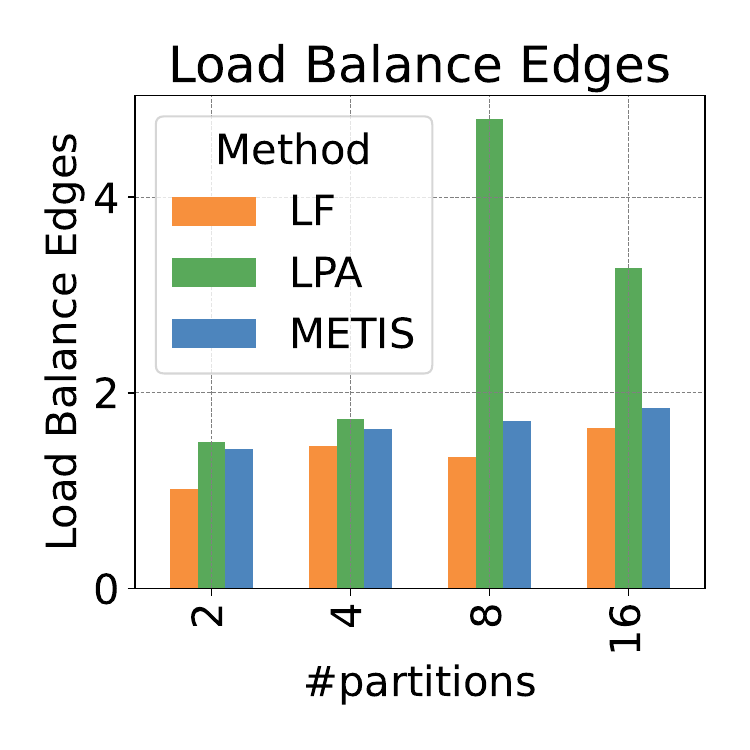}
    \label{fig:fusion}
  \end{subfigure}
  \begin{subfigure}[b]{0.32\linewidth}
    \includegraphics[width=\linewidth]{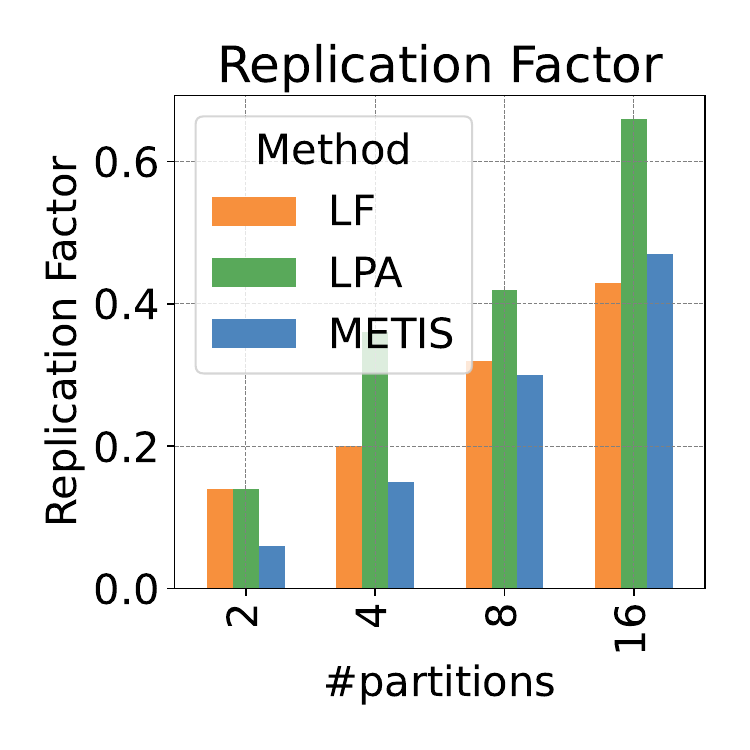}
    \label{fig:fusion}
  \end{subfigure}
  \caption{Comparison of subgraph quality on Arxiv dataset
  }
  \label{fig:partitioning_comparison}
\end{figure}
\vspace{-20pt}
\begin{enumerate}
    \item Edge cuts percentage:
\begin{equation}
\tau=\frac{\sum_{i=1}^k \Gamma\left(V_i, \bar{V}_i\right)}{m}
\end{equation}
which is the sum of edge cuts between each partition $i$ and other partitions $\Gamma\left(V_i, \bar{V}_i\right)$ divided by total number of edges $m$ in the graph. Lower edge cuts represent better partition quality.
    \item Number of connected components for each partition, which is the number of subgraphs of each partition in which each pair of nodes is connected by a path. 
    \item Number of isolated nodes for each partition, which is the number of nodes that are not connected to any other nodes.
    \item Load balance of nodes:
\begin{equation}
\rho=\frac{\max _{i=1, \ldots, k}\left|P_i\right|}{|P_{average}|}
\label{load_balance}
\end{equation}
    where $|P_{average}|=\frac{n}{k}$ is the expected number of nodes for each partition in the ideal situation, and $\max _{i=1, \ldots, k}\left|P_i\right|$ is the maximum number of nodes from $k$ partitions. A lower load balance of nodes represents better partition quality.
    
    \item Load balance of edges: The same formula as for load balance of nodes where $|P_{average}| = \frac{m}{k}$ is the expected number of edges for each partition in the ideal situation, and $\max _{i=1, \ldots, k}\left|P_i\right|$ is the maximum number of edges from $k$ partitions. A lower load balance of edges represents better partition quality.
    
    \item Replication factor:
\begin{equation}
\mathrm{RF}=\frac{1}{n} \sum_{i \in k}|P_i(v)|
\end{equation}
where $n$ is the total number of nodes in the graph, and $P_i(v)$ is the total number of replicas of vertices in each partition.
\end{enumerate}

\Cref{fig:partitioning_comparison} shows the evaluation results of the metrics on the Arxiv dataset, comparing different partitioning methods over different numbers of partitions. 
The results show that our method excels in minimizing the number of connected components and isolated nodes, ensuring that each partition contains only one connected component and no isolated nodes. In contrast, both LPA and METIS result in multiple connected components and numerous isolated nodes.

In terms of edge cuts and replication factor, our method does not show a significant improvement over other methods when considering 2 to 8 partitions. This is to be expected since the primary goal of our method is not to reduce these factors. However, at 16 partitions, our method performs better than others. This improvement can be attributed to the increase in the number of connected components and isolated nodes in other methods, which negatively affects these factors.

\begin{figure}[htbp]
  \centering
  \begin{subfigure}[b]{0.32\linewidth}
    \includegraphics[width=\linewidth]{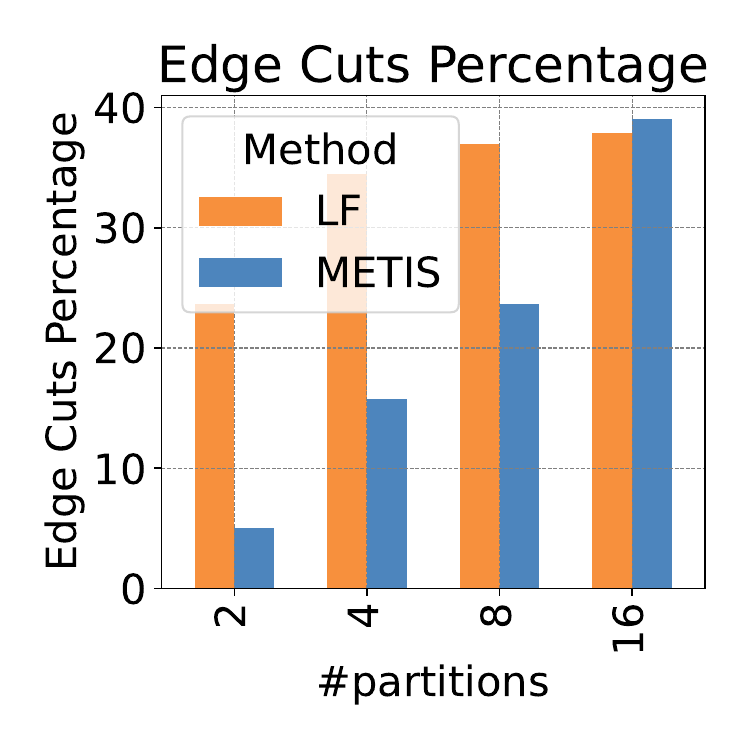}
    \label{fig:leiden}
  \end{subfigure}
  \begin{subfigure}[b]{0.32\linewidth}
    \includegraphics[width=\linewidth]{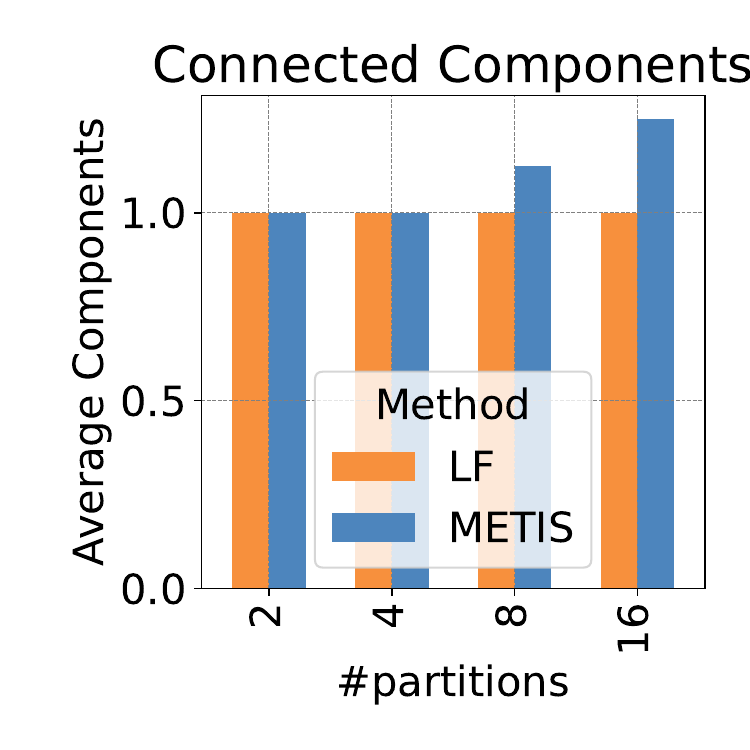}
    \label{fig:fusion}
  \end{subfigure}
  \begin{subfigure}[b]{0.32\linewidth}
    \includegraphics[width=\linewidth]{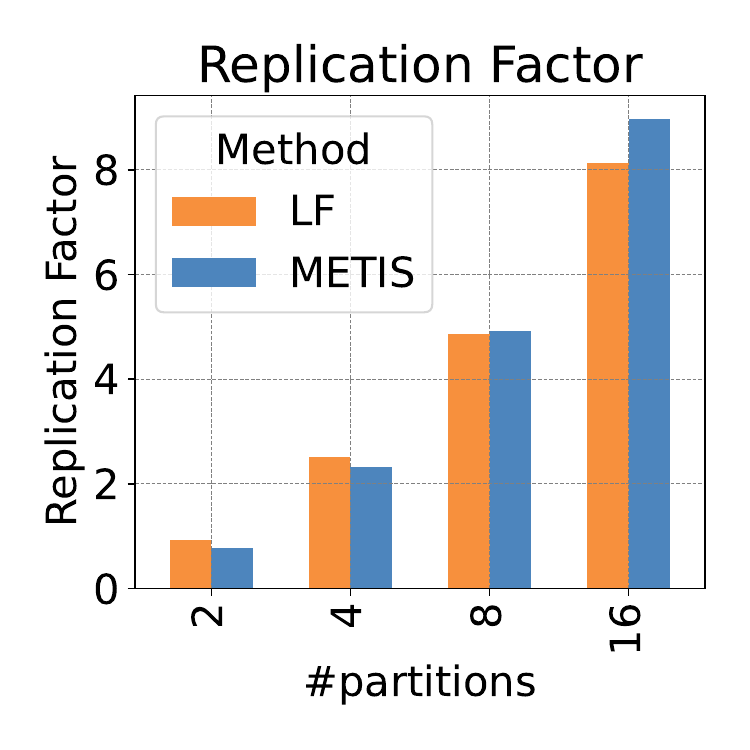}
    \label{fig:fusion}
  \end{subfigure}
  \vspace{-0.7cm}
  \caption{Comparison on Proteins dataset
  }
  \label{fig:partitioning_comparison_proteins}
\end{figure}
\Cref{fig:partitioning_comparison_proteins} shows the results of some metrics on the Proteins dataset. Unlike Arxiv, the Proteins graph is extremely dense with an average node degree of 597, which is 43 times higher than in Arxiv. 
Therefore, the edge-cut percentage and the replication factor are relatively high, but LF performs relatively better on 16 partitions and wins METIS. 
Regarding the number of components, once the number of partitions exceeds 4, METIS fails to achieve a single component per partition, while LF remains successful up to 16 partitions.
\subsection{Quality Comparison}
\label{sec:quality}

We evaluated the overall quality of our solution when applied to downstream tasks by following a specific procedure.
When creating subgraphs based on partition information, we considered two methods: one that ignores edges between partitions (\i.e., inner nodes only) and another that preserves these edges by replicating nodes. These methods will be referred to as \textit{Inner} and \textit{Repli}, respectively. Our goal is to compare the quality of these two approaches.

We train a GCN or GraphSAGE model separately for each partition and obtain the embeddings that are finally combined to train an MLP classifier for the classification task.

\Cref{tab:arxiv_gcn} shows the accuracy comparison of GCN on the Arxiv dataset with \textit{Inner} and \textit{Repli}, for multi-class prediction, from 2 to 16 partitions, compared to the LPA\cite{spark} and METIS\cite{metis} partitioning methods, and \Cref{tab:arxiv_sage} shows the corresponding results for GraphSAGE. 

\begin{figure}[H]
    \centering
    \begin{subfigure}[t]{0.45\textwidth}
        \centering
        \begin{tabular}{ l  c | c | c | c }
        \toprule[1pt]
        \multirow{2}{*}{Method} & \multicolumn{4}{c}{Accuracy (\%)}\\
        \cmidrule(l){2-5}
         & 2 & 4 & 8 & 16 \\
        \midrule
        LPA Inner   & 68.99 & 66.38 & 63.07 & 59.61 \\
        LPA Repli  & 69.60 & 69.57 & 67.97 & 65.62 \\
        \midrule
        METIS Inner & \textbf{69.59} & 68.46 & 65.68 & 60.90 \\
        METIS Repli & 70.32 & 69.86 & 68.95 & 66.70 \\
        \midrule
        Our LF Inner & 69.33 & \textbf{69.09} & \textbf{66.73} & \textbf{65.11} \\
        Our LF Repli  & \textbf{70.34} & \textbf{70.05} & \textbf{69.22} & \textbf{68.19} \\
        \bottomrule[1pt]
        \end{tabular}
        \vspace{0.2cm}
        \caption{Accuracy Comparison of GCN on Arxiv Dataset}
        \label{tab:arxiv_gcn}
    \end{subfigure}
    \hfill
    \begin{subfigure}[t]{0.45\textwidth}
        \centering
        \begin{tabular}{ l  c | c | c | c }
        \toprule[1pt]
        \multirow{2}{*}{Method} & \multicolumn{4}{c}{Accuracy (\%)}\\
        \cmidrule(l){2-5}
         & 2 & 4 & 8 & 16 \\
        \midrule
        LPA Inner & 69.33 & 67.86 & 64.45 & 62.11 \\
        LPA Repli & 69.86 & 68.52 & 67.37 & 62.63 \\
        \midrule
        METIS Inner  & 69.90 & 68.14 & 67.41 & 62.98 \\
        METIS Repli  & 70.22 & 68.54 & 67.29 & 64.25 \\
        \midrule
        Our LF Inner & \textbf{70.63} & \textbf{70.90} & \textbf{68.57} & \textbf{67.58} \\
        Our LF Repli  & 70.48 & 70.46 & \textbf{69.42} & \textbf{68.36} \\
        \bottomrule[1pt]
        \end{tabular}
        \vspace{0.2cm}
        \caption{Accuracy Comparison of SAGE on Arxiv Dataset}
        \label{tab:arxiv_sage}
    \end{subfigure}
    \caption{Accuracy Comparison of different methods on Arxiv Dataset}
\end{figure}
\vspace{-20pt}
In particular, our LF partitioning method significantly improves the quality compared to the METIS and LPA partitioning methods, for both GCN and SAGE algorithms. For GCN on 16 partitions, LF improves METIS by 6.9\% for the \textit{Inner} method and by 2.2\% for the \textit{Repli} method.
It is important to note that LF achieves almost the highest quality possible which is an accuracy of 71\% in a centralized environment. For 16 partitions, the accuracy of the LF method is only 4\% lower than that of the centralized solution, while training remains fully localized with low communication costs. 

Our method also outperforms for both \textit{Inner} and \textit{Repli}. It should be noted that for all methods, the accuracy of \textit{Repli} is higher than that of \textit{Inner}, which is obvious. In addition, compared to the significant accuracy improvement that GCN brings to \textit{Repli} (for example, for LF 16 partitions, the accuracy is improved by 3\%), the improvement for GraphSAGE is not so much (about 1\%).  The reason may be that GraphSAGE uses a neighbor sampling strategy, so the loss of boundary neighbors has less impact on the model.

We now report quality results for the denser Proteins dataset.
Due to its very high density, \textit{Repli} method would replicate too many nodes and increase the training time beyond acceptable limits, thus we only consider the \textit{Inner} method. \Cref{tab:accuracy_comparison_proteins} shows the ROC-AUC results of SAGE model.
\vspace{-10pt}
\begin{table}[H]
\centering
\begin{tabular}{l c c c c}
\toprule[1pt]
\multirow{2}{*}{Method} & \multicolumn{4}{c}{ROC-AUC (\%)}\\
\cmidrule(l){2-5}
 & 2 & 4 & 8 & 16 \\
\midrule
METIS Inner  & 75.48  & 67.53  & 46.45  & 44.80  \\
\midrule
Our LF Inner & 75.21 & 65.13  & \textbf{52.94}  & \textbf{49.38}  \\
\bottomrule[1pt]
\end{tabular}
\vspace{0.3cm}
\caption{Accuracy Comparison of SAGE on Proteins Dataset}
\label{tab:accuracy_comparison_proteins}
\end{table}
\vspace{-20pt}

We can see that for 8 and 16 partitions LF's accuracy is more than 10\% higher than METIS. This may be because METIS partitions have more than one component. In addition, compared to Arxiv, the accuracy of Proteins drops more when the number of partitions is higher (compared to 76\% in centralized training). This may be because we lose more cut edges since the Proteins graph is extremely dense. 

\vspace{-10pt}
\subsection{Speed Analysis}
\label{speed}
\begin{figure}[H]
\hspace{0.5cm}
\begin{minipage}{0.45\textwidth}
\centering
\begin{tabular}{lcccc}
\toprule[1pt]
\multirow{2}{*}{Method} & \multicolumn{4}{c}{Partitioning time (s)} \\
\cmidrule(l){2-5}
                        & 2    & 4    & 8    & 16   \\
\midrule
LPA                     & 71.0 & 104.5 & 173.2 & 327.6 \\
METIS                   & 3.0  & 3.1   & 3.1   & 3.6   \\
Ours (LF)               & 2.1  & 2.0   & 1.8   & 1.7   \\ 
\bottomrule[1pt]
\end{tabular}
\captionof{table}{Partitioning time comparison on Arxiv dataset across different methods and partitioning numbers.}
\label{tab:partitioning_time}
\end{minipage}
\hfill
\begin{minipage}{0.45\textwidth}
\centering
\includegraphics[width=0.9\textwidth]{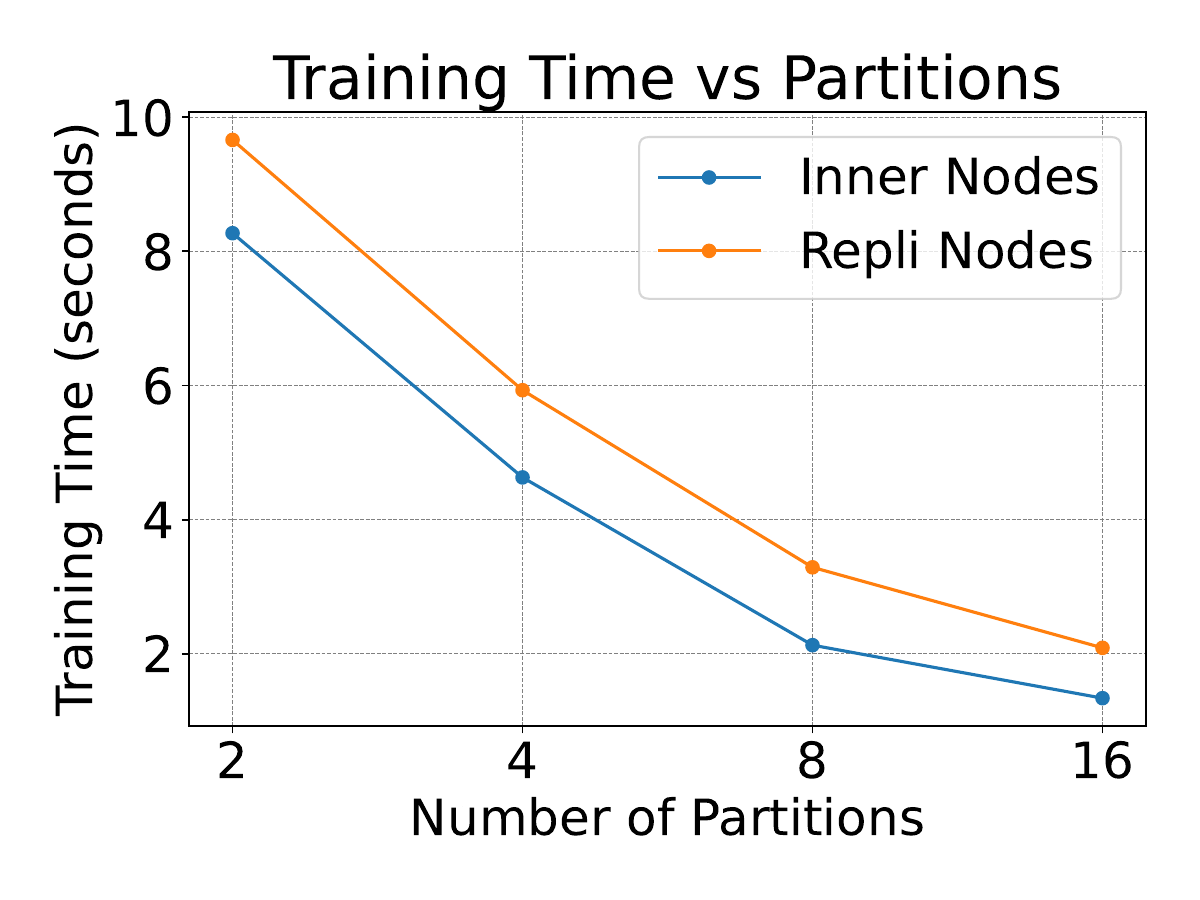}
\caption{Training time of LF on Arxiv using GCN}
\label{fig:training_time}
\end{minipage}
\end{figure}

\Cref{tab:partitioning_time} shows the partitioning time of different partitioning methods on the Arxiv dataset. 
Note that for our LF, there is 11.5s of preprocessing time to find communities using Leiden's library\cite{leiden}. Once we obtain the communities, they can be stored and loaded for further partitioning. 
Another point is that LF is faster when the number of partitions is larger. This is because LF is an iterative greedy algorithm. For example, two partitions can be considered as obtained by continuing to merge from four partitions.
\Cref{fig:training_time} shows the longest training time of all subgraphs obtained by our LF algorithm using the GCN model on the Arxiv dataset. It can be seen that increasing the number of partitions dramatically reduces the training time, while for synchronized distributed frameworks such as DGL\cite{dgl} and PBG\cite{pbg}, the training time does not decrease much due to numerous communications as discussed in Spark Local\cite{spark}. Also, for each partition with \textit{Repli}, the training time increases only a little compared to \textit{Inner}, while the accuracy is much higher as shown in \Cref{sec:quality}.
\vspace{-20pt}

\vspace{-10pt}
\subsection{Impact of Our Fusion Method on Other Partitioning Methods}
To further evaluate the benefits of our fusion method, we compared its performance on different partitioning methods. We report the results of our fusion method applied to METIS, LPA, and Leiden for 16 partitions on the GCN model, focusing on partitioning time, edge cuts percentage, and accuracy on the Arxiv dataset.
On \Cref{tab:fusion_time_edge_cuts} we named each method with a "+F" suffix, which stands for "fusion".

\begin{table}[ht]
    \centering
    \begin{tabular}{lccc}
        \toprule
        Method & Time(s) & Edge cuts before F(\%)  & Edge cuts after F(\%)  \\
        \midrule
        METIS+F & 4.8 & 25.4 & 25.1 \\ 
        LPA+F & 6.6 & 28.0 & 27.0 \\ 
        Leiden+F & 1.7 & - &  23.7 \\
        \bottomrule
    \end{tabular}
    \vspace{0.3cm}
    \caption{Partitioning time(s) and Edge Cuts(\%) for 16 partitions on Arxiv}
    \label{tab:fusion_time_edge_cuts}
\end{table}

\vspace{-20pt}

We observe that our fusion method reduces the percentage of edge cuts for both METIS and LPA partitioning methods, resulting in improved partition quality.
Regarding the fusion time, we note that the fusion process is 2.2 times faster when applied to Leiden compared to METIS (and 3.9 times faster compared to LPA). This is because Leiden inherently guarantees connected communities, whereas for METIS and LPA, we need to additionally identify each connected component.

\Cref{tab:fusion_accuracy}  shows the accuracy results for GCN model on Arxiv dataset. Comparing to \Cref{tab:arxiv_gcn}, we can observe that our fusion method highly improved the accuracy results for both METIS and LPA partitioning methods,  \textit{Inner} results is comparable to the Leiden Fusion, while LF yieds better results for \textit{Repli}. The combination of our Leiden + Fusion method proves its efficiency and effectiveness. 
\begin{table}[ht]
    \centering
    \begin{tabular}{lccccc}
        \toprule
        Method & METIS & METIS+F & LPA & LPA+F & Leiden+F  \\
        \midrule
        Inner & 60.90 & \textbf{65.75} & 59.61 & 64.51 & 65.11 \\
        Repli & 66.70 & 67.60 & 65.62 & 66.85 & \textbf{68.19} \\
        \bottomrule
    \end{tabular}
    \vspace{0.3cm}
    \caption{Accuracy results (\%) for GCN 16 partitions}
    \label{tab:fusion_accuracy}
\end{table}
\vspace{-25pt}
\section{Conclusion}

Current partitioning methods and distributed frameworks face two major challenges in effectively training GNNs 
that hinder the handling of large networks: 1) the need for continuous communication in synchronized distributed frameworks to access information from other machines, and 2) the inability to ensure that subgraphs remain connected components without isolated nodes. To address these issues, we introduce Leiden-Fusion, a novel partitioning method designed for large-scale graph training with minimal communication.
We made the following contributions: (i) For any initially connected graph, our novel partitioning method ensures that each partition is a single densely connected component with no isolated nodes, facilitating effective GNN training. (ii) By adopting a local training strategy without communication, we significantly reduced the training time while maintaining most of the embedding quality. This approach demonstrates that high training efficiency is achievable for GNNs without sacrificing accuracy, enabling more scalable and efficient distributed learning on very large graphs. In future work, we plan to extend our method to handle graphs with multiple components and isolated nodes, and to evaluate its accuracy and efficiency on graphs with different size densities.

\section{Acknowledgement}

This work is funded by the SCAI (Sorbonne Center for Artificial Intelligence) at Sorbonne University, France.

\bibliographystyle{splncs04}

\end{document}